\documentclass[runningheads]{llncs}
\usepackage{graphicx}
\begin{document}
\title{FusionNet: Incorporating Shape and Texture for Abnormality Detection in 3D Abdominal CT Scans}
\titlerunning{FusionNet: Incorporating Shape and Texture for Abnormality Detection }
%
%
\author{Fengze Liu\inst{1} \and
Yuyin Zhou\inst{1} \and Elliot Fishman\inst{2} \and
Alan Yuille\inst{1}}
\authorrunning{F. Liu et al.}
%
\institute{ The Johns Hopkins University, Baltimore, MD 21218, USA \and
 The Johns Hopkins University School of Medicine, Baltimore, MD 21287, USA
}
\maketitle              
\begin{abstract}
 Automatic abnormality detection in abdominal CT scans can help doctors improve the accuracy and efficiency in diagnosis. In this paper we aim at detecting pancreatic ductal adenocarcinoma (PDAC), the most common pancreatic cancer. Taking the fact that the existence of tumor can affect both the shape and the texture of pancreas, we design a system to extract the shape and texture feature at the same time for detecting PDAC. In this paper we propose a two-stage method for this 3D classification task. First, we segment the pancreas into a binary mask. Second, a FusionNet is proposed to take both the binary mask and CT image as input and perform a binary classification. The optimal architecture of the FusionNet is obtained by searching a pre-defined functional space. We show that the classification results using either shape or texture information are complementary, and by fusing them with the optimized architecture, the performance improves by a large margin. Our method achieves a specificity of 97\% and a sensitivity of 92\% on 200 normal scans and 136 scans with PDAC.

\end{abstract}

\section{Introduction}

Pancreatic cancer is one of the most dangerous type of cancer. In 2019, about 56770 people will be diagnosed with pancreatic cancer, and pancreatic cancer accounts for about 3\% of all cancers in the US and about 7\% of all cancer deaths~\cite{ACS}. The 5-year relative survival rate for all stages of pancreatic cancer is only about 9\%, while it can rise to 34\% if the cancer is detected in an early stage. However, even experienced doctors may miss an early stage cancer because it is small and hard to observe. So developing an reliable automatic system to assist doctors to diagnosis can help decrease the missing rate of patients with early stage of cancer.

This paper is aimed at discriminating normal cases from cases with pancreatic ductal adenocarcinoma (PDAC), the major type of pancreatic cancer accounting for about 85\% of the cases, by checking into the abdominal 3D CT scans. With the development of deep learning in recent years~\cite{NIPS2012}, researchers have made significant progress in automatically segmenting organs like pancreas from CT scans~\cite{roth2015deeporgan,zhou2017fixed,zhu3d}, which is already a hard task due to the irregular shape of pancreas~\cite{zhang2017personalized}. Even though, segmenting the lesion region is an even more challenging task due to the large variation in shape, size and location of the lesion~\cite{zhou2017deep}. And the full annotation for the lesion region requires more expertise and time to obtain. So instead of directly segmenting the lesion region, detecting the patients with PDAC can already help the diagnosis and, more importantly, is more feasible when the annotation is limited. 

We choose to utilize the segmentation mask and CT image for pancreatic abnormality detection, since the segmentation mask can represent the shape while the CT image represents the texture, which are both important for abnormality detection. However we find that the classification results of using only shape and only texture information are quite complementary, which motivates us to combine them in a unified system and thereby can improve the classification outcome. In the natural image domain, how to effectively combine different information has been explored in several different works.~\cite{depth} proposes a fusion network incorporating depth to improve the segmentation.~\cite{marrnet} calculates the normal, depth and silhouette from a single image for better 3D reconstruction. Other works like~\cite{multiview,yuyin} build different networks for different views of the same data and present co-training strategy to enable the models to incorporate different views.

In this paper we develop a two-stage method for this problem. Firstly, a recent state-of-the-art segmentation network~\cite{Yu2018RecurrentST} is used to segment the pancreas and then tested on all the data to get the prediction mask for pancreas. Secondly, the CT image is fed into a deep discriminator together with the prediction mask. The discriminator is employed to extract information from both the image and segmentation mask for abnormality classification. We optimize the architecture of the discriminator by searching from a functional space, which includes functions with different fusion strategies. Unlike~\cite{Zhu2018MultiScaleCS} that needs full annotation for the lesion, our method only requires annotation masks for the pancreas region on cases without PDAC in the first stage, and image-level labels indicating abnormality in the second stage. Other works like~\cite{Chen2018,fengze2018} make use of the information from either the prediction mask or CT image for classification. We show in the experiments that these two kinds of information are complementary to each other and the combination can improve the classification result by a large margin.

We test our framework on 200 normal and 136 abnormal (with PDAC) CT scans. We report a 92\% sensitivity and 97\% specificity, {\em i.e.} missing 11 out of 136 abnormal cases with 6 false alarms out of 200 normal cases. Compared with using only single branch, our method improves the result by more than 5\% in specificity and 10\% in sensitivity.

\begin{figure*}[!t]

\includegraphics[width=12cm]{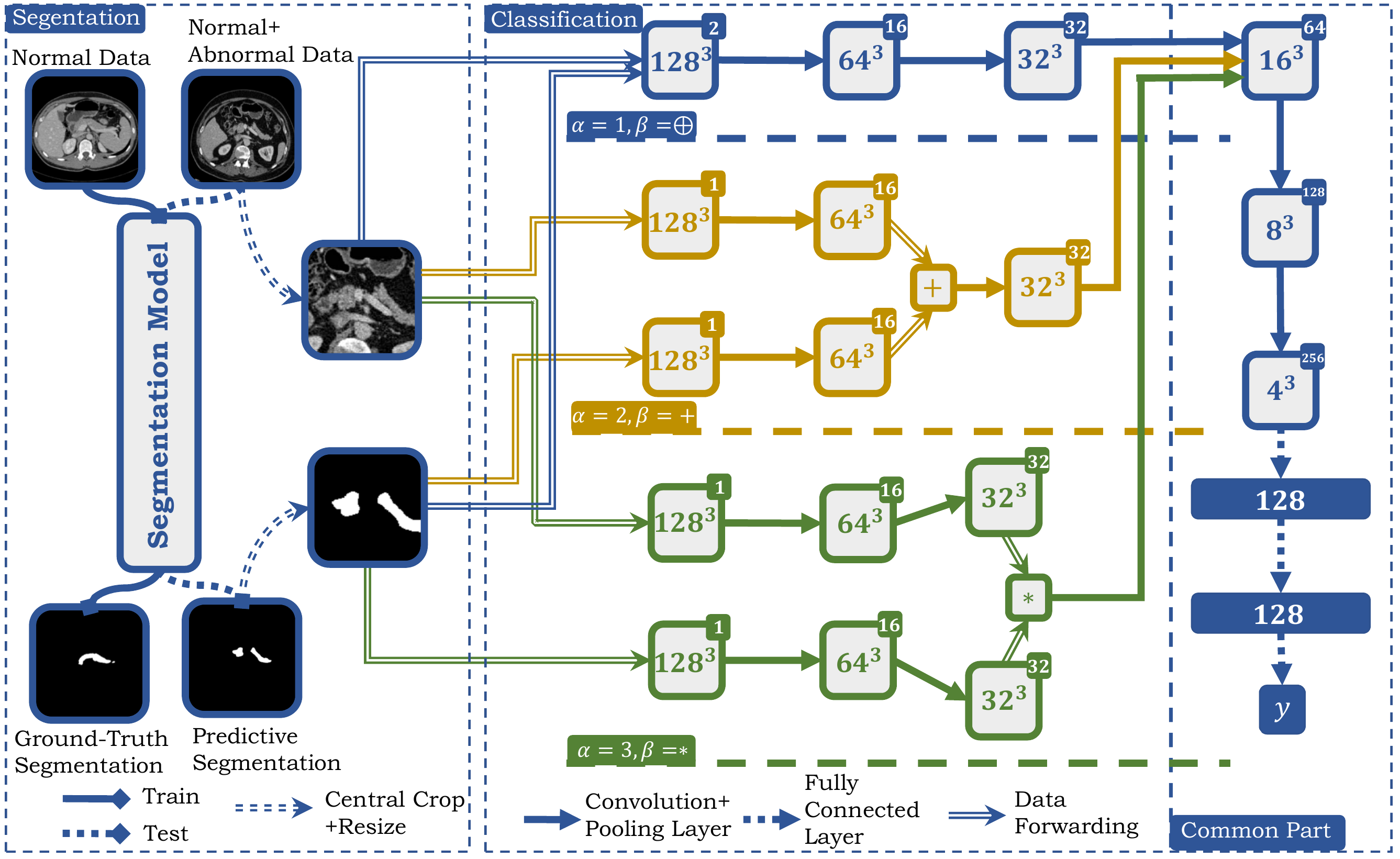}

\caption{
   The pipeline of our framework. In stage $1$, a segmentation network is trained using the normal data. Then the segmentation network is tested on both normal and abnormal data. The 3D mask and image are cropped and scaled as the input of second stage. At the right side, we show the examples of fusion model using different $\alpha, \beta$. Note that these three models share the same architecture after layer $3$ because $\alpha<=3$ in the examples but they do not share the weights. Each convolution layer uses a set of $3\times3\times3$ kernels, and each pooling layer uses $2\times2\times2$ kernels with a stride of 2. Batch normalization and ReLU activation are used after all these layers. 
}
\label{Fig:pipeline}

\end{figure*}

\section{Fusion Network for detecting PDAC}
\subsection{The Overall Framework}
The CT scan $\bf{X}\in\mathcal{X}$ is a volume of size $L\times W\times H$, where $L\rm{,}W\rm{,}H$ represents the length, width and height of the volume respectively. Typically, a CT scan is of size $512\times512\times H$, where $H$ is the number of slices along the axial axis. Each element in the volume indicates the Hounsfield Unit (HU) at a certain position. Our goal is to learn a discriminative function $f(\bf{X})\in\{0,1\}$, where 1 indicates PDAC and 0 otherwise. 

Directly learning the function $f(\cdot)$ is feasible but not optimal. Because the high dimensionality and rich texture information in the CT image can easily make the model overfit, especially when the number of training data is limited.~\cite{fengze2018} introduces a constraint by segmenting the pancreas first and learn $f(g(\bf{X}))$, where $g(\cdot)$ is a segmentation function to get a binary mask of pancreas $\bf{S}$. However this will result in loss of texture information since $g(\bf{X})$ is only a binary mask. In order to fully exploit both shape and texture information we consider learning $$f(g(\bf{X}),\bf{X}),$$ which takes both the segmentation mask and image as input. The major problem here is how to design the function $f(\cdot)$ so that it can well extract shape information from $g(\bf{X})$ and texture information from $\bf{X}$ and combine them for the classification task. Our idea is to define a functional space representing a set of different fusion strategies and the optimal architecture is obtained by searching that functional space. Given a normal CT dataset $\mathcal{X}_1=\{\bf{(X},\bf{Y})\}$, where the annotation for pancreas $\bf{Y}$ is available, and $\mathcal{X}_2=\{(\bf{X},z)\}$ which contains both normal and abnormal cases with only image-level label $\bf{z}$ indicating the abnormality, we split our framework into two stages. First we train a segmentation function $g(\cdot)$ on $\mathcal{X}_1$ and test it on $\mathcal{X}_2$, then the prediction masks together with CT images on $\mathcal{X}_2$ become the input for the second stage to train a classification function $f(\cdot)$. We will introduce each stage in detail in the following sections.

\subsection{The Segmentation Stage}
This stage is necessary in the framework for getting the segmentation mask which will provide shape information in the second stage. Since the focus in this paper is how to combine $g(\bf{X})$ and $\bf{X}$ in $f(\cdot)$, and also the two stages are executed separately, so the form of $g(\cdot)$ is out of range of this study and will be investigated in the future. In this paper we choose a recent stat-of-the-art segmentation framework~\cite{Yu2018RecurrentST} for $g$. Since $g(\cdot)$ is a 2D-based method so we need to concatenate the output of different slices to reconstruct the 3D volume like in~\cite{bridge}. We train the segmentation algorithm on $\mathcal{X}_1$ and test it on $\mathcal{X}_2$. After that, we crop out the region-of-interest(ROI) from both CT image and prediction mask, defined as the cube bounding box covering all foreground voxels in the prediction mask and padded by 20 voxels in each dimension. Then the cropped regions are resampled to $128\times128\times128$ volumes. We denote the predictive mask after cropping and resampling as $\bf{\hat{S}}=g(\bf{X})$.

\subsection{The Classification Stage}
The two branches of the input represent different information. The image domain contains rich texture information, while the binary mask can indicate shape of the target object. Directly concatenating them in the very first layer is an intuitive way but may not be optimal. To explore the optimal fusion strategy, we start from a base model with $L=6$ convolution layers similar with 3D VNet~\cite{3dv}, followed by two fully connected layers, as shown in Figure \ref{Fig:pipeline}. Then a functional space for different architectures is defined as $\{(\alpha,\beta)|\alpha\in\{1,2,...,L\},\beta\in\{+,*,\oplus\}\}$, where $\alpha$ indicates at which layer to fuse and $\beta$ indicates how to fuse. Here $\oplus$ represents concatenation. See also in Figure \ref{Fig:pipeline} for specific examples for different combination of $\alpha, \beta$. We formulate each fusion function in the functional space $f_{\alpha\beta}(\cdot)$ as following.

$$ f_{\alpha\beta}({\bf{S}},{\bf{X}};w) = f_{\alpha:L} (\beta(f_{1:\alpha}({\bf{S}};w_{1:\alpha}^1),f_{1:\alpha}({\bf{X}};w_{1:\alpha}^2));w_{\alpha:L}). $$

Here $f_{1:\alpha}(\cdot)$ is the first $\alpha$ convolution layers of the base model while $f_{\alpha:L}(\cdot)$ is the remaining layers. $w=\{w_{1:\alpha}^1,w_{1:\alpha}^2,w_{\alpha:L}\}$ is the parameters to learn. The feature maps of two branches after the first $\alpha$ layers are fused using operation $\beta(\cdot)$ as $\beta(f_{1:\alpha}({\bf{S}};w_{1:\alpha}^1),f_{1:\alpha}({\bf{X}};w_{1:\alpha}^2))$, and then fed into $f_{\alpha:L}(\cdot)$. The idea of this design is to alleviate the effect of changing the model structure but only focus on finding the best way to combine two different input.

Once given $\alpha,\beta$, we learn $w$ by optimizing a weighted cross-entropy loss

$$L=-\lambda\log{p^z}-(1-\lambda)\log(1-p)^{1-z},$$
where $p=f_{\alpha\beta}(\hat{S},X;w)$. The output of $f_{\alpha\beta}(\cdot)$ is activated by a sigmoid function so that $p\in[0,1]$. $z\in\{0,1\}$ is the label for a CT scan indicating whether this study suffers from PDAC. We set $\lambda=0.7$ for balancing the class difference during training.

\section{Experiments}

In this section we test our two-stage framework on our dataset containing 3D abdominal CT scans with both patients with and without PDAC. We compare our method with other method using single source input and also show the result of different fusion architectures. We report the sensitivity(SEN), specificity(SPEC), ROC AUC Score(AUC) and F1 Score(F1) to evaluate the classification model.

\vspace{-0.2cm}
\subsection{Dataset and Settings}
We collect the dataset with the help of the radiologists. There are 300 normal cases and 136 biopsy proven PDAC cases. 100 out of 300 normal cases have voxel-wise annotations for pancreas (denoted as set $\mathcal{X}_1$), and the remaining 200 normal cases as well as the 136 PDAC cases only have image-level labels, {\em i.e.}, abnormal/normal (denoted as set $\mathcal{X}_2$). In the first stage, we train the segmentation network on $\mathcal{X}_1$ and test it on $\mathcal{X}_2$. In the second stage, $\mathcal{X}_2$ is randomly split into four folds for cross-validation, where each fold contains 50 normal and 34 abnormal cases and the fusion network is trained on three of the folds and tested on the remaining one.

For the first stage, we follow the instruction of~\cite{Yu2018RecurrentST} to train a segmentation network. For the second stage, we apply grid search on $\alpha$ and $\beta$, {\em i.e.} we choose for every pair of $(\alpha,\beta)\in\{(\alpha,\beta)|\alpha\in\{1,2,...L\},\beta\in\{+,*,\oplus\}\}$. In our case, $L=6$, so there are $3L=18$ different architectures in total in the search space. After setting $\alpha$ and $\beta$, for training $f_{\alpha\beta}(\cdot)$, we use stochastic gradient descent(SGD) with batch size of $4$. The learning rate is set to 0.01 with exponential decay rate 0.9997. We also perform data augmentation on both the CT image and prediction mask by slightly rotating $0^\circ,\pm 10^\circ$ along three axes individually (27 possibilities) to prevent from overfitting, since the number of training data is very limited. For each pair of $\alpha,\beta$, the model is trained for $10\rm{,}000$ iterations, which takes about 1.5 hours on a NVIDIA TITAN RTX(24GB) GPU.

\begin{figure}[!t]
\begin{center}

    \includegraphics[width=12.5cm]{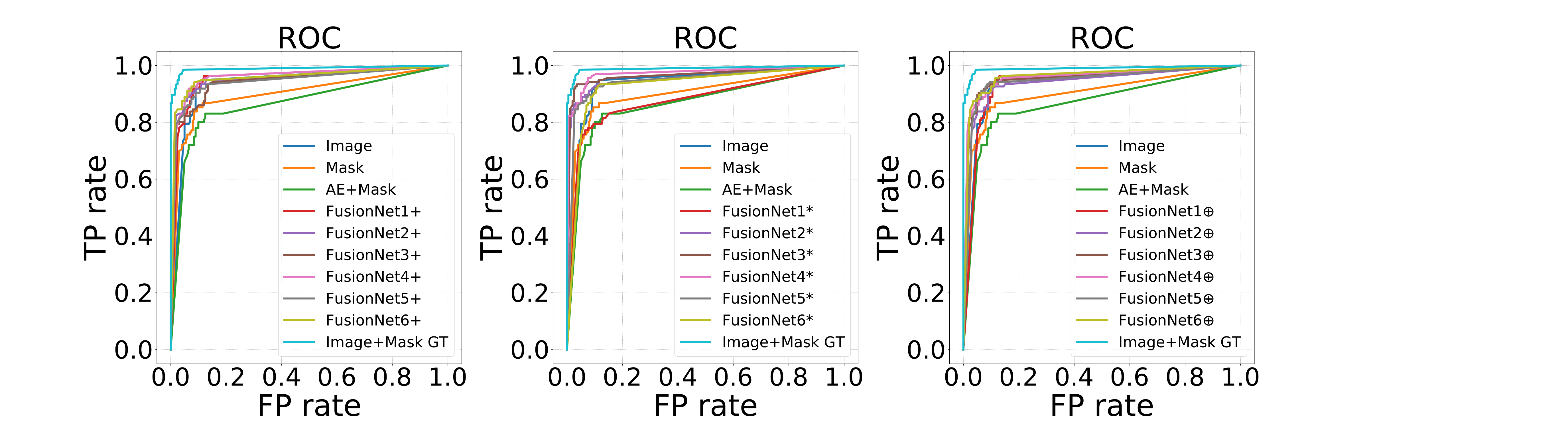}
\end{center}

\caption{
   ROC curves for comparison of different fusion strategies. {\bf{Left}}: fused by $+$. {\bf{Mid}}: fused by $*$.  {\bf{Right}}: fused by $\oplus$. The Image, Mask and AE+Mask are the baseline methods without fusing. The Image+Mask GT is the pseudo upper bound of the fusing. 
}
\label{Fig:ROC}

\end{figure}

\begin{table}
    \centering
    \setlength{\tabcolsep}{3.5mm}
    \begin{tabular}{l c c c c }
\hline

\multicolumn{1}{l|}{}  & SEN & SPEC & AUC & F1 \\
\hline\hline
\multicolumn{1}{l|}{AE+Mask~\cite{fengze2018}} & 77.94&	91.00&	89.04&   \multicolumn{1}{c}{81.54} \\
\multicolumn{1}{l|}{Mask} & 82.35&	91.50&	92.94&   \multicolumn{1}{c}{84.53} \\
\multicolumn{1}{l|}{Image} & 83.09&	92.00&	95.95&  \multicolumn{1}{c}{85.28} \\

\hline

\multicolumn{1}{l|}{Naive Fusion} & 83.09 & 95.50 &	97.17 &  \multicolumn{1}{c} {87.60 }\\

\hline
\multicolumn{1}{l|}{FusionNet3*(Ours)} & {\bf{92.65}}&	{\bf{97.00}}&	{\bf{97.72}}&  \multicolumn{1}{c}{\bf{94.03}} \\

\hline
\multicolumn{1}{l|}{Mask+Image GT} & 94.12&	97.50&	99.53&	\multicolumn{1}{c}{95.17}\\

\hline
\\
\end{tabular}

    \caption{Comparison between our method and baseline methods on the sensitivity(SEN), specificity(SPEC), area under the curve(AUC) and F1 score(F1). FusionNet3* achieves the best result, indicating the best way to fuse is to multiply two branches in the third layer.}
    \label{tab:result}

\end{table}

\subsection{Primary Results}
We compare our method with~\cite{fengze2018} which utilizes the feature from pre-trained auto-encoder for classification (AE+Mask). We also compare with the base model using either the CT image (Image) or prediction mask (Mask) as input. Our fusion model has the same network structure with the base model after the fusing point for fair comparison. The best result is achieved when fusing the two branches in third layer with multiplication operation. The result is summarized in Table \ref{tab:result}. The ROC curves of different models are shown in Figure \ref{Fig:ROC}. Image+Mask GT indicates the strategy that if either one of the two methods (Image and Mask) correctly classifies the case, then we treat this case as correctly classified. This can be the upper bound of merging because it fuses the result based on the ground-truth label. The large improvement in the upper bound result shows that the information provided by the CT image and prediction mask for abnormality detection are quite complementary to each other, which proves the necessity of combining them together. Naive Fusion is done by taking the average of output from Mask and Image and only shows limited improvement. This fact further validates the efficacy of our proposed FusionNet.

\vspace{-0.2cm}
\subsection{Analysis and Discussion}
\paragraph{Single Branch Comparison:}
From Table \ref{tab:result} we can see the comparison  among Image, Mask and AE+Mask which all use only one branch of information. Using only the image works the best, which indicates the importance of texture for detecting PDAC. For the other two methods using only shape information, directly training a discriminator achieves better results, showing that the constraint of auto-encoder can harm the classification performance. 

\begin{figure}[!t]
\begin{center}
    \includegraphics[width=12.5cm]{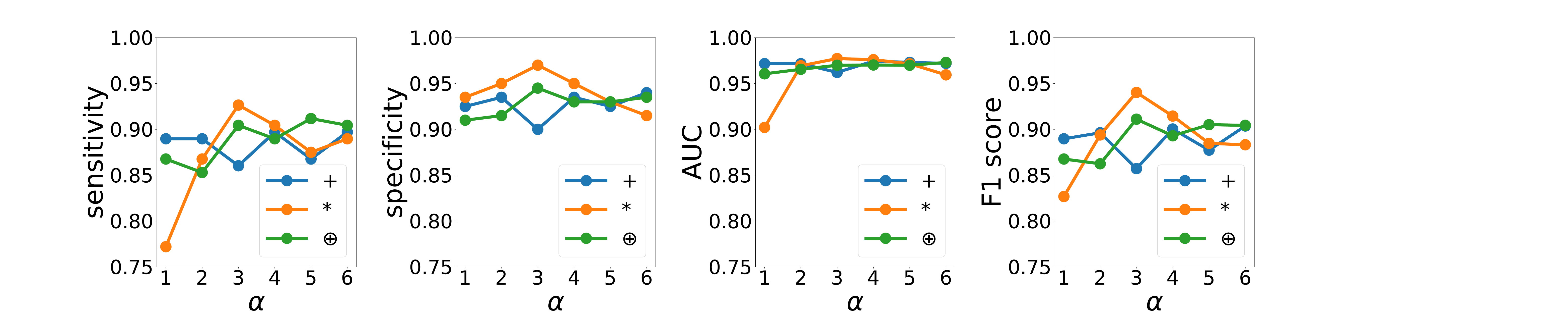}
\end{center}

\caption{
   Comparison on the sensitivity, specificity, AUC and F1 score between different fusion architectures. 
}
\label{Fig:compare}

\end{figure}

\begin{table*}
    \centering
    \setlength{\tabcolsep}{2mm}
    \begin{tabular}{l c c c c c c c}
\hline

\multicolumn{2}{c|}{$\alpha$} & 1 & 2 & 3 & 4 & 5 & 6 \\
\hline\hline
\multicolumn{1}{l|}{} &\multicolumn{1}{c|}{F1} & 88.97&	89.63&	85.71& 90.04&	87.73&	90.37	 \\
\multicolumn{1}{l|}{$\beta=+$}&\multicolumn{1}{c|}{\# Para} & 3.99M&	3.99M&	4.01M& 4.10M&	4.45M&	5.86M	\\
\multicolumn{1}{l|}{}&\multicolumn{1}{c|}{FLOPs}  & 7.99M&	7.99M&	8.04M& 8.21M&	8.92M&	11.74M	\\

\hline\hline
\multicolumn{1}{l|}{} &\multicolumn{1}{c|}{F1} & 82.68&	89.39&	94.03& 91.45&	88.48&	88.32	 \\
\multicolumn{1}{l|}{$\beta=*$}&\multicolumn{1}{c|}{\# Para}  & 3.99M&	3.99M&	4.01M& 4.10M&	4.45M&	5.86M	\\
\multicolumn{1}{l|}{}&\multicolumn{1}{c|}{FLOPs}  & 7.99M&	7.99M&	8.04M& 8.21M&	8.92M&	11.74M	\\

\hline\hline
\multicolumn{1}{l|}{} &\multicolumn{1}{c|}{F1} & 86.76& 86.25&	91.11& 89.30&90.51&	90.44	 \\
\multicolumn{1}{l|}{$\beta=\oplus$}&\multicolumn{1}{c|}{\# Para}  & 3.99M&	4.01M&	4.07M& 4.32M&	5.34M&	7.96M	\\
\multicolumn{1}{l|}{}&\multicolumn{1}{c|}{FLOPs}  & 7.99M&	8.02M&	8.15M& 8.66M&	10.69M&	15.94M	\\

\hline
\\
\end{tabular}
    \caption{Comparison between different fusion architectures on the F1 scores, number of parameters and floating-point operations(FLOPs). As $\alpha$ increases, the two branches fuse at more latter layer, so the parameters number also increases. However the best performance is achieved when $\alpha=3$.}
    \label{tab:para}
\end{table*}
\vspace{-0.5cm}
\paragraph{Fusion Comparison:}
The result of fusing at different layers with different operations is as shown in Table \ref{tab:para} and Figure \ref{Fig:compare}. First of all, almost all the fusion models can perform better than the single branch model, which proves the advantages of fusing shape and texture. Table \ref{tab:para} shows the number of parameters and floating-point operations for each model to show how the size of model affects the classification result. We can see as $\alpha$ increases, the size of model increases, but the classification result does not always improve correspondingly. For the $+$ fusion operation, the performance is better when fusing at the earlier or later layers of the network. For the $*$ and $\oplus$ operation, however, fusing at the middle layer of the network shows better performance. The best result is obtained when fusing at the third layer with $*$ operation.

\section{Conclusion}
In this paper we propose a FusionNet which combines shape and texture information from the segmentation mask and CT image for detecting PDAC. Compared with using only single source of information, using both shape and texture information improves the performance by a large margin. We also explore the best network structure for fusing these two branches together by searching from a functional space, which is to multiply the feature map of two branches in the middle of the network. We report a 92\% sensitivity and a 97\% specificity by doing 4-fold cross-validation on 200 normal patients and 138 patients with PDAC.

%
%
%
\bibliographystyle{splncs04}
\bibliography{reference.bib}

\end{document}